\documentclass{article}

\usepackage[preprint]{corl_2026} 
\usepackage[table]{xcolor}
\usepackage[numbers]{natbib}
\usepackage{multicol}
\usepackage{algorithm}
\usepackage[noend]{algpseudocode}
\usepackage{amsmath}
\usepackage{amsfonts}
\usepackage{amssymb}
\usepackage{xspace}
\usepackage{graphicx}
\usepackage{float}
\usepackage{color}
\usepackage{subcaption}

\usepackage{wrapfig}
\usepackage{booktabs}
\usepackage{multirow}
\usepackage[table]{xcolor}
\definecolor{mygray}{RGB}{230,230,230}

\newif\ifcomments
\commentstrue

\ifcomments
    \newcommand{\caelan}[1]{{\textcolor[rgb]{0.0,0.0,1.0}{[Caelan: {\it #1}]}}}
    \newcommand{\ajay}[1]{{\textcolor[rgb]{1.0,0.0,1.0}{[Ajay: {\it #1}]}}}
    \newcommand{\shuo}[1]{\textcolor[rgb]{0.2,0.3,0.7}{[Shuo: #1]}}
    \newcommand{\todo}[1]{\textcolor[rgb]{1.0,0.0,0.0}{[TODO: #1]}}
    \newcommand{\alfie}[1]{{\color{orange}(alfie: #1)}}
\else
    \newcommand{\caelan}[1]{}
    \newcommand{\ajay}[1]{}
    \newcommand{\shuo}[1]{}
    \newcommand{\todo}[1]{}
    \newcommand{\alfie}[1]{}
\fi

\newcommand{\ours}{Human2Any\xspace} 

\title{Human2Any: Human-to-Robot Transfer via Constraint-Aware Compositional Planning}

\author{
  \textbf{Shuo Cheng}\textsuperscript{1},
  \textbf{Chuye Zhang}\textsuperscript{1},
  \textbf{Alfred Cueva}\textsuperscript{1}, \\[0.5em]
  \textbf{Caelan Garrett}\textsuperscript{2,*},
  \textbf{Ajay Mandlekar}\textsuperscript{2,*},
  \textbf{Danfei Xu}\textsuperscript{1} \\[0.5em]
  \textsuperscript{1}Georgia Institute of Technology \quad
  \textsuperscript{2}NVIDIA Corporation
}

\begin{document}
\maketitle


\begin{abstract}
    Human videos are a scalable source of supervision for robot manipulation, as they are abundant and naturally capture rich object interactions. However, transferring human demonstrations to robots remains challenging due to embodiment mismatch, scene variation, and robot-specific feasibility constraints. We present \textbf{\ours}, a framework for learning reusable object-centric interaction priors from human videos without requiring real-world robot demonstrations in the target task contexts. \ours represents manipulation through object--object interaction motion, capturing task-relevant scene changes while abstracting away embodiment-specific details. It composes learned interaction priors with robot-side feasibility reasoning and motion planning, allowing the same human-derived knowledge to adapt to different embodiments, scene geometries, and task contexts. We validate \ours across diverse manipulation settings, including real-world experiments on a Franka tabletop setup and an RBY-1 humanoid mobile robot, demonstrating robust interaction-centric manipulation without real-world robot training data. Project website: \url{https://human2any.github.io/}.

    
\end{abstract}

\def\thefootnote{$\ast$}\footnotetext{Equal Contribution.}\def\thefootnote{\arabic{footnote}}

\begin{figure}[H]
    \centering
    \includegraphics[width=\linewidth]{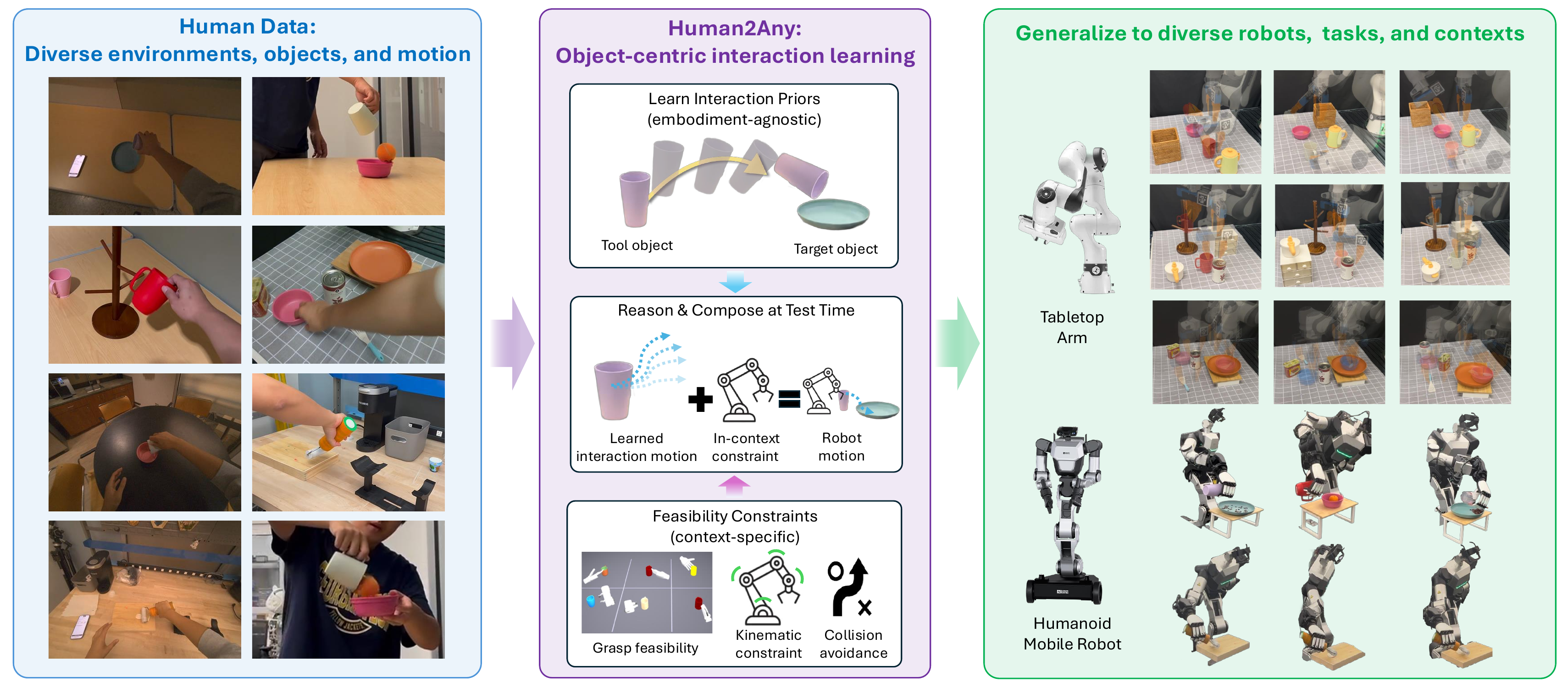}
    \caption{\textbf{Human-to-robot transfer with Human2Any.} {\ours} learns object-centric interaction priors from human videos and adapts them to diverse robot embodiments, tasks, and contexts.}
    \label{fig:teaser}
    \vspace{-0.75em}
\end{figure}


\section{Introduction}

Learning manipulation systems that can generate feasible and purposeful motions for solving everyday tasks remains a fundamental yet challenging problem~\cite{firoozi2024foundation,liu2023libero,Zhao_2025,trilbmteam2025carefulexaminationlargebehavior}. While {behavior cloning (BC)} has achieved remarkable progress in recent years~\cite{chi2023diffusionpolicy, robomimic2021, rt22023arxiv, florence2021implicit, Ze2024DP3}, its generalization to diverse environments---such as varying object geometries, poses, and scene layouts---heavily depends on both the quality and quantity of training data~\cite{robomimic2021, saxena2025mimiclabs}. Unfortunately, the collection of high-quality, embodiment-aligned demonstrations, typically obtained through robot teleoperation~\cite{open_x_embodiment_rt_x_2023, khazatsky2024droid}, is time-consuming and costly, which significantly limits scalability.

In contrast, \emph{learning from human videos} has emerged as a promising paradigm for teaching robots at scale~\cite{kareer2025egomimic, xu2024flow, bharadhwaj2024track2act}. 
Unlike robot demonstrations, human videos are abundant and diverse, but their most transferable signal is not the human body motion itself; rather, it is the \emph{object-centric interaction structure} underlying the task.
Human videos capture rich manipulation strategies~\cite{hoque2025egodex, bharadhwaj2023generalizable}: they show how objects change state \emph{because of} purposeful interactions---reach, grasp, insert, pour---across long horizons and varied objects, layouts, and styles. This diversity is ideal for learning \emph{compositional} skills, but directly imitating human motion is mismatched to robot embodiment. Trajectory-level reproducibility is inherently context-dependent: a motion that works in one scene may be infeasible on the robot due to reachability, joint limits, or collisions. 
As the layout changes, previously infeasible motions may become feasible, and vice versa. Thus, the key is not to copy a particular trajectory, but to extract structure that remains stable across embodiments and contexts.

This raises a central challenge: \emph{how can we harness the breadth of human videos to learn transferable interaction models that can be adapted to robot-specific constraints?} We tackle this problem through the lens of \textbf{object-centric motion generation}. Our key insight is that what transfers is the \emph{interaction structure}: which objects participate, when contacts form and break, and how object states evolve. Accordingly, we represent the manipulation process as a {factor graph} that explicitly disentangles \textit{agent--object} and \textit{object--object} interactions (Fig.~\ref{fig:teaser}). The object--object factors capture agent-agnostic motion distributions learned from human videos, while the agent--object factors and feasibility terms account for embodiment-specific constraints such as grasping, reachability, and collision avoidance. At test time, we sample and refine graph variables under the robot's in-context constraints to produce robot-feasible interaction sequences that preserve the goal-achieving structure observed in video. This formulation scales the acquisition of diverse object-centric motion models while retaining structure-aware adaptability across embodiments and task scenarios, unifying compositional reasoning and constraint satisfaction within a sampling-and-scoring procedure.

\textbf{Our main contributions are summarized as follows:}
\vspace{1mm}
\newline \noindent $\bullet$ We formulate manipulation as an \textit{object-centric factor graph} that separates human-transferable \textit{object--object} interaction factors from robot-specific \textit{agent--object} feasibility factors, enabling motion learning from human videos with test-time robot adaptation.
\vspace{1mm}
\newline \noindent $\bullet$ We represent skills as compositions of object-centric motion models and develop a foundation-model-based pipeline~\cite{ravi2024sam, xiao2025spatialtrackerv2, wen2024foundationpose} to extract object motions from unstructured human videos.
\vspace{1mm}
\newline \noindent $\bullet$ We introduce a constraint-aware sampler that scores and resamples candidate motion compositions under in-context constraints to produce robot-feasible motions.
\vspace{1mm}
\newline \noindent $\bullet$ We validate the approach across multiple robot embodiments and long-horizon tasks, demonstrating generalization, cross-embodiment adaptability, and planning efficiency.
\section{Related Work}

\noindent\textbf{Human-video-based manipulation learning.}
Human videos provide a scalable source of supervision for robot manipulation because they capture diverse object interactions without requiring robot teleoperation. Prior work learns from human demonstrations through visual imitation, retargeting, affordance prediction, or intermediate object-centric signals such as 2D/3D object flow and point tracks~\cite{xu2024flow,bharadhwaj2024track2act,yuan2024general,noh20253d,liu2025egozero,shi2025zeromimic,kareer2025egomimic,wang2023mimicplay,bharadhwaj2023generalizable}. Other works co-train robot policies with human videos and robot data, often using human videos to improve visual representations or rewards~\cite{punamiya2025egobridge,nair2022r3m,radosavovic2023real,ma2022vip}. Retargeting-based methods explicitly address embodiment mismatch by mapping human motion to robot actions, while flow- or trajectory-conditioned methods use object motion as a transfer interface. However, these methods typically predict robot actions directly or rely on learned translation policies, making it difficult to enforce robot-specific feasibility, scene geometry, and contact constraints at deployment. In contrast, \ours extracts embodiment-agnostic object-object motion from human videos and resolves embodiment- and scene-specific constraints at test time.

\noindent\textbf{Object-centric and interaction-centric manipulation.}
Object-centric representations have been widely used to improve manipulation generalization across object instances, configurations, and scenes. Prior methods learn affordances, grasping strategies, object flows, motion fields, or post-grasp object trajectories to guide manipulation~\cite{xu2021deep,fang2023anygrasp,wang2022dexgraspnet,eppner2021acronym,song2020grasping,Eisner-RSS-22,hsu2025spot,seita2023toolflownet,zhang2023flowbot++,li2024flowbothd,yin2025object,su2025motion,li2025novaflow}. These approaches often decompose manipulation into grasp selection and motion generation, but many follow a cascaded workflow in which a grasp or intermediate motion representation is chosen before checking downstream feasibility. Recent work jointly optimizes grasp and trajectory from video demonstrations~\cite{dong2025joint} or decomposes manipulation into alignment and interaction stages using robot-specific data~\cite{dreczkowski2025learning}. \ours instead learns object--object interaction motion from human videos and composes it with robot-specific grasping, feasibility checking, and motion planning, enabling test-time adaptation to the current embodiment and scene context.

\noindent\textbf{Compositional planning and guided sampling.}
Task-and-motion planning decomposes long-horizon manipulation into symbolic decisions and continuous motions~\cite{kaelbling2011hierarchical,garrett2021integrated,garrett2020pddlstream}. Learning has been integrated into TAMP through learned skills, samplers, or constraints to improve scalability and generalization~\cite{cheng2023league,silver2022learning,mandlekar2023hitltamp,cheng2023nod,mandalika2019generalized,chitnis2016guided,chitnis2019learning,wang2021learning,ortiz2022structured,fang2024dimsam,kumar2026open}. Meanwhile, diffusion models have become effective trajectory generators for robot learning and long-horizon composition~\cite{ho2020denoising,song2020denoising,song2020score,chi2023diffusionpolicy,Ze2024DP3,mishra2023generative,mishra2024generative,luo2025generative,luo2024potential,clark2025you}. However, independently sampled motion components may be individually plausible but jointly infeasible under grasp, kinematic, collision, and scene constraints. \ours addresses this through constraint-aware test-time steering: sampled interaction and grasp motions are assembled into robot trajectories, scored with in-context feasibility checks, and progressively resampled toward executable compositions.%

\section{Learning Interaction Priors from Video}
\label{sec:learning_priors}

\begin{figure}[t]
    \centering
    \includegraphics[width=\linewidth]{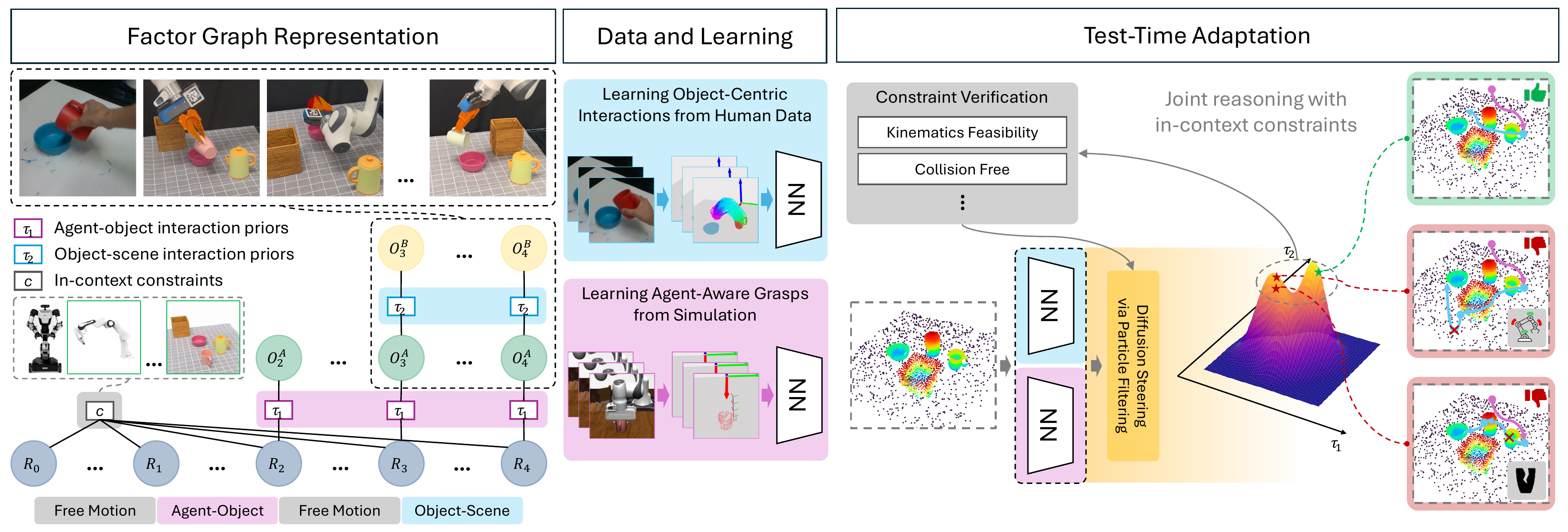}
    \caption{\textbf{\ours overview.}
    Given human demonstration videos, \ours extracts object-centric interaction motion between tool and target objects and learns object--object interaction priors that are independent of the human embodiment. Separately, robot-specific agent--object priors model how a particular robot embodiment grasps and controls tool objects. At deployment, these learned priors are composed under the current robot, scene, and task constraints to generate executable manipulation trajectories.}
    \label{fig:pipeline}
    \vspace{-1em}
\end{figure}

Human videos provide rich object-interaction demonstrations, but their motions are not directly executable by robots due to embodiment mismatch and scene-dependent constraints. We treat each video as a tool object interacting with a target object, e.g., a cup tilting toward a bowl or a utensil being placed on a bowl, and extract object geometry and relative object motion while discarding human-specific arm and hand motion. As shown in Fig.~\ref{fig:pipeline}, \ours learns object--object interaction priors from human videos and robot-specific agent--object priors from embodiment-specific grasp data, then composes them with robot- and scene-specific feasibility reasoning at deployment.


\noindent\textbf{Interaction-Centric Task Representation.} We consider tasks specified by a high-level skeleton $\mathcal{S}=[s^1,\ldots,s^K]$~\cite{garrett2021integrated}, where each phase $s^k=\langle a^k,\mathcal{O}^k\rangle$ denotes an interaction type $a^k$ and the involved objects $\mathcal{O}^k$. There are three interaction types: free-space motion, agent--object interaction, and object--object interaction. Free-space motion connects interaction phases; agent--object interaction establishes robot-specific contact with a tool object; and object--object interaction describes how a tool object moves relative to a target object.

The key transferable representation is the object--object interaction trajectory $\tau^{\mathrm{rel}}=\{\mathbf{T}^{\mathrm{rel}}(h)\}_{h=0}^{H-1}$, where $\mathbf{T}^{\mathrm{rel}}(h)\in\mathrm{SE}(3)$ denotes the tool-object pose relative to the target object at time $h$. This representation captures task-relevant interaction motion while removing the human embodiment from the supervision signal. In this work, we focus on prehensile manipulation, where the robot maintains a rigid attachment to the tool object during interaction. This allows robot execution to be recovered from object--object interaction motion and a grasp pose. 

\noindent\textbf{Object--Object Priors from Human Videos.} For each processed demonstration, we assume access to the tool and target object point clouds $\mathbf{P}^A,\mathbf{P}^B \in \mathbb{R}^{N \times 3}$ and the corresponding object-centric relative trajectory $\tau^{\mathrm{rel}}=\{\mathbf{T}^{\mathrm{rel}}_h\}_{h=0}^{H-1}$. In our implementation, these are estimated from RGB-D videos using off-the-shelf segmentation and tracking tools~\cite{ravi2024sam,xiao2025spatialtrackerv2}; details are provided in the appendix. Given tracked 3D keypoints $\xi^{3d}\in\mathbb{R}^{H\times M\times 3}$, we estimate $\tau^{\mathrm{rel}}$ using the Kabsch algorithm~\cite{lawrence2019purely} with RANSAC~\cite{fischler1981random}. We trim each trajectory to frames near the interaction segment using a task-dependent distance threshold from the final interaction pose, approximately $0.15$--$0.2$ m across tasks. This yields $\mathcal{D}_{\mathrm{os}}=\{(\mathbf{P}^A_i,\mathbf{P}^B_i,\tau^{\mathrm{rel}}_i)\}_{i=1}^{|\mathcal{D}_{\mathrm{os}}|}$ for training object--object interaction samplers.




\noindent\textbf{Robot-Specific Agent--Object Priors.}
Object--object priors specify how objects should interact, but the robot must still choose an embodiment-specific way to grasp and control the tool object. We therefore train an agent--object prior for each robot embodiment using simulated grasp data. Each sample contains a tool-object point cloud $\mathbf{P}^{A}$ and a robot end-effector trajectory $\tau^{e}$ for reaching and grasping the object, forming  $\mathcal{D}_{\mathrm{ro}}=\{(\mathbf{P}^A_i,\tau^e_i)\}_{i=1}^{|\mathcal{D}_{\mathrm{ro}}|}$. The resulting prior $p_{\theta_{\mathrm{ro}}}(\tau^{e}\mid \mathbf{P}^{A})$ captures embodiment-specific grasping behavior while remaining separate from the object--object interaction prior. This separation allows object-level interaction knowledge learned from human videos to be reused across different robot embodiments, while robot-specific execution is handled through the agent--object prior and test-time feasibility reasoning.

\noindent\textbf{Diffusion Trajectory Prior Training.}
We instantiate object--object and agent--object priors as conditional diffusion models over pose trajectories. Given a clean trajectory $\mathbf{x}_0$, the forward process generates noisy samples $\mathbf{x}_t=\sqrt{\bar{\alpha}_t}\mathbf{x}_0+\sqrt{1-\bar{\alpha}_t}\boldsymbol{\epsilon}$, where $\boldsymbol{\epsilon}\sim\mathcal{N}(\mathbf{0},\mathbf{I})$. The model predicts the injected noise by minimizing $\mathcal{L}(\theta)=\mathbb{E}_{t,\mathbf{x}_0,\boldsymbol{\epsilon}}[\|\boldsymbol{\epsilon}-\boldsymbol{\epsilon}_{\theta}(\mathbf{x}_t,t,\tilde{c})\|_2^2]$, where $\tilde{c}$ denotes the corresponding point-cloud conditioning input. For object--object priors, $\tilde{c}=(\mathbf{P}^{A},\mathbf{P}^{B})$; for agent--object priors, $\tilde{c}=\mathbf{P}^{A}$. We apply rigid transformations to both point clouds and trajectories during training to improve robustness to object poses and scene layouts. Additional details are provided in the Appendix. 

\section{Constraint-Aware Composition}
\label{sec:composition}

At deployment, the robot observes a context $c$ containing the robot model, scene point cloud, and task skeleton. Each learned prior can generate diverse candidate motion segments, but independently plausible segments may not compose into an executable robot trajectory. We therefore formulate test-time execution as constraint-aware composition: jointly selecting motion segments that remain likely under the learned priors while satisfying robot- and scene-specific constraints.

\subsection{Test-Time Context and Trajectory Assembly}

Let $\mathbf{x}^{1:K}=\{\mathbf{x}^{1},\ldots,\mathbf{x}^{K}\}$ denote sampled motion segments for the $K$ phases in the task skeleton. We define an assembly operator $\Gamma(\mathbf{x}^{1:K},c)$ that converts these segments into a full robot trajectory and evaluates feasibility under the current context. For agent--object segments, $\Gamma$ directly uses the sampled end-effector trajectory. For object--object segments, $\Gamma$ converts the sampled relative object trajectory into end-effector targets using the grasp transform provided by the preceding agent--object segment. Under the rigid grasp assumption, this conversion is $\mathbf{T}^{e}(h)=\mathbf{T}^{\mathrm{rel}}(h)\mathbf{T}^{e}(0)$, where $\mathbf{T}^{e}(0)$ is the end-effector pose after grasping the tool object. The assembled trajectory is then checked for kinematic feasibility, collision avoidance, and task-specific constraints between segments.

This feasibility depends on both the robot embodiment and the current scene. For example, in \textsc{PourInBowl}, a sampled cup-to-bowl trajectory may be plausible at the object level but infeasible if the wrist collides with nearby obstacles or the pouring pose lies outside the robot workspace. Another candidate with a different grasp or pouring direction may remain feasible. Thus, object--object priors alone are insufficient; sampled interactions must be composed with robot- and scene-specific constraints at test time.

\noindent\textbf{Composition by Rejection Sampling.}
A natural composition strategy is to sample each segment independently from its learned prior and accept the composition only if $\Gamma(\mathbf{x}^{1:K},c)$ is executable. Let $\mathcal{E}$ denote the event that the assembled trajectory satisfies all constraints, and let $q(c)=\Pr(\mathcal{E}=1\mid c)$ be the feasible mass under this independent proposal. The expected number of trials is $\mathbb{E}[N_{\mathrm{trial}}]=1/q(c)$. This strategy becomes inefficient when feasible compositions occupy only a small region of the joint sample space, which often occurs under tight grasp, kinematic, and collision constraints. 

\subsection{Constraint-Aware Steering}

\begin{wrapfigure}{r}{0.48\linewidth}
    \centering
    \vspace{-1.em}
    \includegraphics[width=0.98\linewidth]{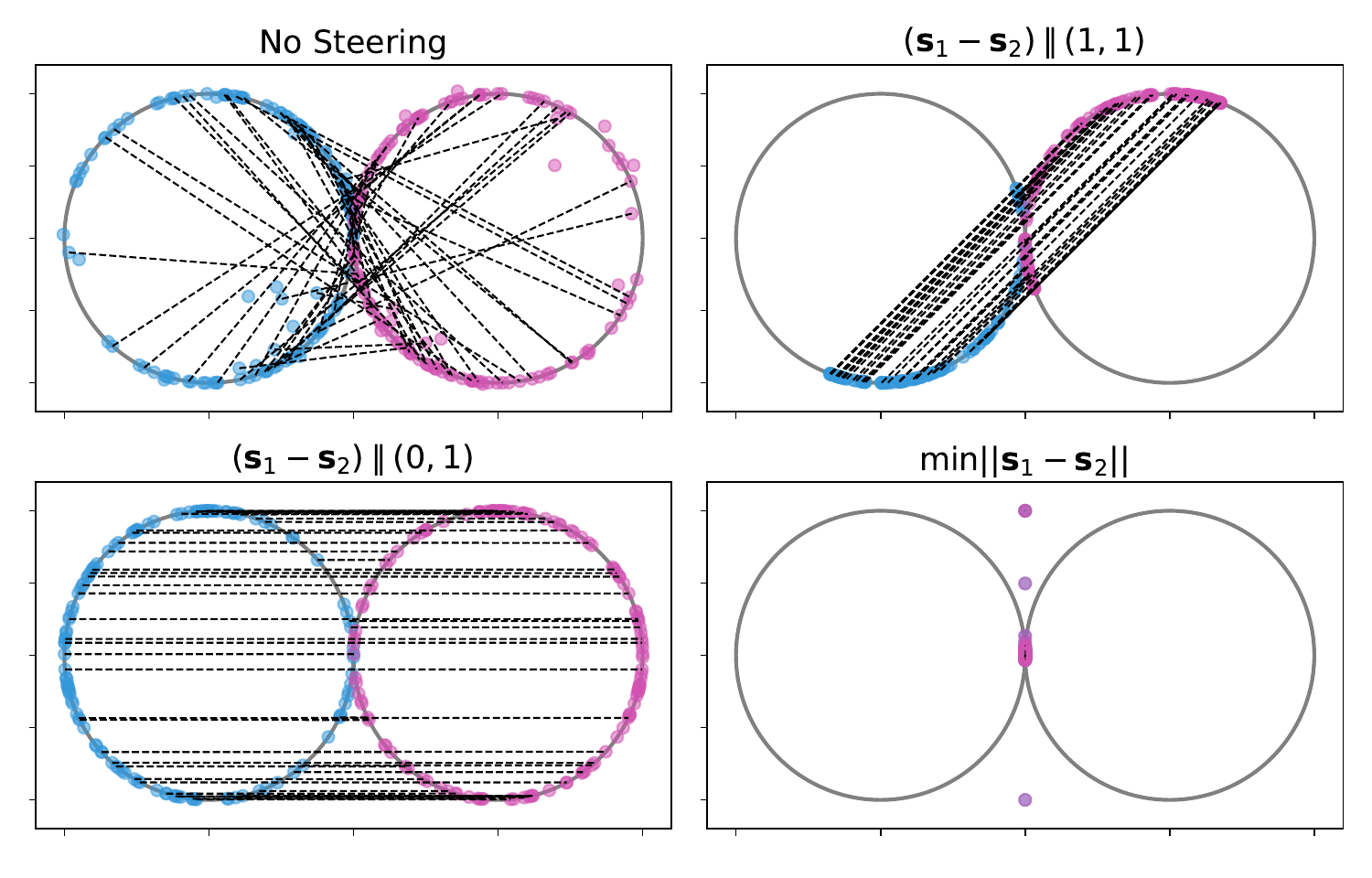}
    \caption{\textbf{Toy illustration.}
    Independent samplers produce locally plausible samples, but feasible joint compositions occupy only a small subset of the product space. Steering uses feasibility scores during denoising to concentrate particles around valid compositions.}
    \label{fig:toy}
    \vspace{-1.em}
\end{wrapfigure}

Fig.~\ref{fig:toy} illustrates why independent sampling can be inefficient. Each sampler assigns high probability to locally plausible samples, shown as the two circles, but only a small subset of their joint combinations satisfies the global composition constraint, such as being parallel to a reference direction. Rejection sampling checks feasibility only after complete samples are generated, wasting computation on incompatible combinations. In contrast, our method guides generation toward feasible compositions during denoising.

We formulate the desired distribution as $p(\mathbf{x}^{1:K}\mid c,\mathcal{E}=1)\propto p(\mathcal{E}=1\mid \mathbf{x}^{1:K},c)\prod_{k=1}^{K}p_{\theta_k}(\mathbf{x}^{k}\mid \tilde{c}^{k})$, where $\tilde{c}^{k}$ is the segment-level conditioning input and $p_{\theta_k}$ is the diffusion prior for phase $k$. The feasibility likelihood is $p(\mathcal{E}=1\mid \mathbf{x}^{1:K},c)\propto \exp(\beta \mathcal{S}(\Gamma(\mathbf{x}^{1:K},c),c))$, where $\mathcal{S}$ scores the assembled trajectory under the current in-context constraints, and $\beta$ controls the strength of steering.

We approximate this posterior using particle filtering over the joint reverse diffusion process. At denoising step $t$, we maintain $M$ particles $\{\mathbf{x}^{1:K}_{t,i}\}_{i=1}^{M}$. For each particle, we estimate the clean segment trajectories using Tweedie's formula, $\hat{\mathbf{x}}^{\,k}_{0,i}=\frac{1}{\sqrt{\bar{\alpha}_t}}(\mathbf{x}^{k}_{t,i}-\sqrt{1-\bar{\alpha}_t}\boldsymbol{\epsilon}_{\theta_k}(\mathbf{x}^{k}_{t,i},t,\tilde{c}^{k}))$. The denoised segment estimates are assembled by $\Gamma$ and scored with weights $w_{t,i}\propto\exp(\beta\mathcal{S}(\Gamma(\hat{\mathbf{x}}^{\,1:K}_{0,i},c),c))$. Particles are resampled according to these weights before the next reverse-diffusion step.

The score $\mathcal{S}$ is soft rather than binary: it can incorporate continuous measures such as IK residuals, collision margins, and motion-planning feasibility. Therefore, particles need not be fully executable at early denoising steps; steering gradually reallocates computation toward particles whose denoised estimates are closer to satisfying the constraints. The full procedure is provided in the Appendix.

\subsection{Implementation Details}

We implement $\Gamma$ using \textsc{MPlib}~\cite{guo2026mplib} to synthesize free-space motions and verify kinematic and collision constraints directly on observed scene point clouds, avoiding the need for explicit object mesh models during planning. Once a feasible trajectory is found, we execute it with platform-specific controllers: OSC~\cite{khatib2003unified} at $20\,\mathrm{Hz}$ for Franka end-effector targets, and joint-position control for the RBY-1 torso, arms, and dexterous hands, with PD control for mobile-base tracking. Camera, perception, and additional planning details are provided in the appendix.


%
\section{Experiments}

Our experiments evaluate whether \ours enables scalable and transferable manipulation learning from object-centric interaction motion. Specifically, we test six hypotheses. \textbf{H1:} \ours can acquire challenging manipulation skills without embodiment-aligned robot data. \textbf{H2:} \ours generalizes to new task contexts without in-context data. \textbf{H3:} \ours can execute real-world manipulation tasks without real-world robot training data. \textbf{H4:} \ours transfers across substantially different robot embodiments. \textbf{H5:} Constraint-aware steering is critical for efficient and feasible composition. \textbf{H6:} \ours benefits from scaled interaction-motion data.

\subsection{Experimental Setup and Baselines}

\begin{figure}[t]
  \centering
  \includegraphics[width=1\linewidth]{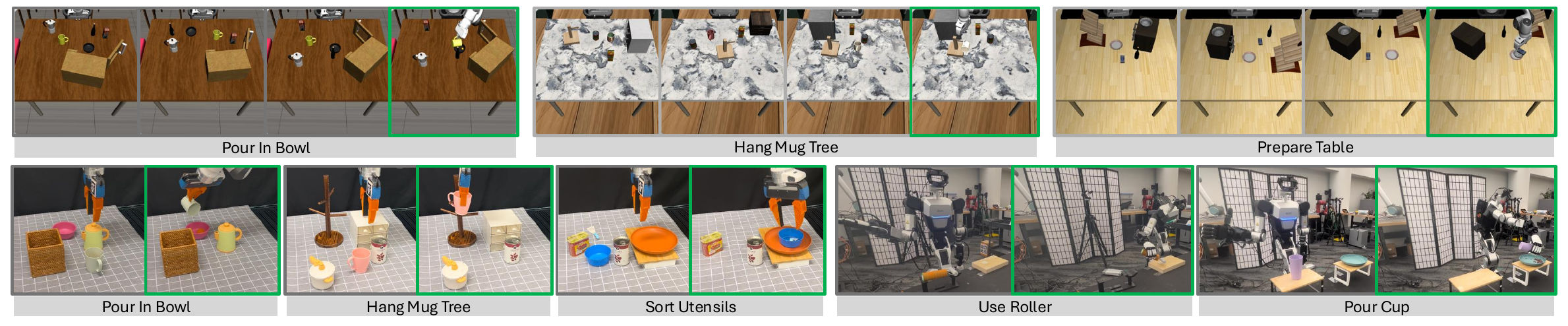}
  \caption{\textbf{Tasks.} We evaluate \ours on a diverse set of simulated and real-world manipulation tasks spanning fine-grained object interactions, pick-and-place, tool use, and long-horizon skill chaining. For each simulation task, we show the initial states of three variants, along with one successful final state highlighted by a green border. These tasks cover diverse object shapes, poses, and scene layouts, and are instantiated on different robot embodiments, including a Franka tabletop setup and an RBY-1 humanoid mobile robot.}
  \label{fig:task_overview}
\end{figure}

\textbf{Simulation tasks.}
We design three MuJoCo-based domains from MimicLab~\cite{todorov2012mujoco,saxena2025mimiclabs}: \textsc{PourInBowl}, \textsc{HangMugTree}, and \textsc{PrepareTable} (Fig.~\ref{fig:task_overview}, top). These tasks evaluate fine-grained motion, long-horizon control, and compositional generalization. Each domain includes three variants with distinct layouts, object arrangements, and interaction contexts. Within each variant, objects are further randomized by approximately $0.1$ m translation and $30^\circ$ rotation. We train on two variants and reserve the third for OOD evaluation. Detailed task descriptions are provided in the appendix.

\textbf{Real-world tasks.}
We evaluate real-world execution without real-world robot training data on two distinct platforms: a Franka tabletop setup and an RBY-1 humanoid with dexterous hands (Fig.~\ref{fig:task_overview}, bottom). For Franka, we use \textsc{PourInBowl}, \textsc{HangMugTree}, and \textsc{SortUtensils}; for RBY-1, we use \textsc{PourCup} and \textsc{UseRoller}. We run ten trials per task with varied object shapes and randomized poses to test robustness across real-world configurations.


\textbf{Baselines.} We compare against DP3~\cite{Ze2024DP3}, an in-domain behavior cloning policy trained with 100 robot demonstrations per task in simulation and 50 real-robot demonstrations per task in real-world experiments; Im2Flow2Act~\cite{xu2024flow}, which predicts 2D object flow from human videos and translates it into robot actions using simulation data; and a Rejection Sampling ablation that uses the same learned components as \ours but disables steering.

\subsection{Main Results: Zero-Shot Composition and Generalization}

We first evaluate \ours on the simulation benchmark in Tab.~\ref{tab:main_compare}, testing challenging skill acquisition without embodiment-aligned robot demonstrations (\textbf{H1}), generalization to new task contexts without in-context data (\textbf{H2}), and the benefit of constraint-aware steering (\textbf{H5}). Each domain contains two in-distribution variants (S1--S2) and one OOD variant with substantially different scene layouts, object arrangements, and interaction contexts.

\begin{table}[H]
\centering
\small
\setlength{\tabcolsep}{3pt}
\resizebox{1\linewidth}{!}{%
\begin{tabular}{l|ccc|ccc|ccc|cc}
\toprule
\multirow{2}{*}{\textbf{Method}} 
& \multicolumn{3}{c|}{\textbf{PourInBowl}} 
& \multicolumn{3}{c|}{\textbf{HangMugTree}} 
& \multicolumn{3}{c|}{\textbf{PrepareTable}} 
& \multicolumn{2}{c}{\textbf{Overall}} \\
\cmidrule(lr){2-4} \cmidrule(lr){5-7} \cmidrule(lr){8-10} \cmidrule(lr){11-12}
& \textbf{S1} & \textbf{S2} & \textbf{OOD}
& \textbf{S1} & \textbf{S2} & \textbf{OOD}
& \textbf{S1} & \textbf{S2} & \textbf{OOD}
& \multicolumn{1}{c}{\textbf{ID}}
& \multicolumn{1}{c}{\textbf{OOD}} \\
\midrule
DP3
& 0.46 & 0.50 & 0.00
& 0.18 & 0.10 & 0.00
& 0.00 & 0.10 & 0.00
& \multicolumn{1}{c}{\cellcolor{mygray}0.22}
& \multicolumn{1}{c}{\cellcolor{mygray}0.00} \\


Im2Flow2Act
& 0.10 & 0.04 & 0.00
& 0.02 & 0.00 & 0.00
& 0.00 & 0.00 & 0.00
& \multicolumn{1}{c}{\cellcolor{mygray}0.03}
& \multicolumn{1}{c}{\cellcolor{mygray}0.00} \\
\midrule
Reject Sampling (No Steering)
& 0.72 & \textbf{0.56} & 0.52
& \textbf{0.72} & 0.42 & 0.40
& 0.46 & \textbf{0.66} & \textbf{0.72}
& \multicolumn{1}{c}{\cellcolor{mygray}0.59}
& \multicolumn{1}{c}{\cellcolor{mygray}0.55} \\

{Ours}
& \textbf{0.86} & \textbf{0.56} & \textbf{0.80}
& \textbf{0.72} & \textbf{0.50} & \textbf{0.60}
& \textbf{0.60} & 0.56 & \textbf{0.72}
& \multicolumn{1}{c}{\cellcolor{mygray}\textbf{0.63}}
& \multicolumn{1}{c}{\cellcolor{mygray}\textbf{0.71}} \\
\bottomrule
\end{tabular}%
}
\caption{\textbf{Simulation benchmark results.} Success rate comparison across three manipulation tasks under in-distribution (ID; S1--S2) and out-of-distribution (OOD) settings. ID and OOD report average success rates over the corresponding settings.}
\label{tab:main_compare}
\vspace{-2.5em}
\end{table}

\textbf{\ours solves challenging tasks without embodiment-aligned data.}
Across the three domains, \ours achieves the strongest overall performance. DP3 performs substantially worse despite using 100 robot demonstrations per task, suggesting that direct action regression struggles with diverse object shapes, poses, layouts, and long-horizon error accumulation. Im2Flow2Act also underperforms, indicating that flow-based action translation does not sufficiently capture embodiment feasibility and 3D scene constraints. These results support \textbf{H1}.


\textbf{\ours generalizes to new contexts without in-context data.}
On OOD variants with unseen layouts and object arrangements, \ours maintains strong performance and outperforms baselines across all domains. This suggests that object--object interaction motion decouples task-relevant interaction structure from specific training layouts, enabling generalization to new poses, shapes, and scene contexts. These results support \textbf{H2}.

\textbf{Constraint-aware steering improves feasible composition.}
Compared with Rejection Sampling, which uses the same learned components but disables steering, \ours achieves more consistent performance and stronger OOD generalization. Thus, jointly steering generation with in-context constraints is more effective than independently sampling and filtering candidates, supporting \textbf{H5}.

\subsection{Real-world Evaluation}

\begin{wraptable}{r}{0.45\linewidth}
\vspace{-1em}
\centering
\scriptsize
\resizebox{\linewidth}{!}{
\begin{tabular}{l|ccc|c}
\toprule
Method & \textsc{PIB} & \textsc{HMT} & \textsc{SU} & Avg. \\
\midrule
DP3 & 4/10 & 4/10 & 0/10 & 0.27 \\
Im2Flow2Act & 1/10 & 2/10 & 0/10 & 0.10 \\
\textbf{Ours} & \textbf{8/10} & \textbf{7/10} & \textbf{9/10} & \textbf{0.80} \\
\bottomrule
\end{tabular}
}
\caption{\textbf{Real-world tabletop results.} Success rate comparison across three real-world tabletop manipulation tasks. Task names are abbreviated using initials.}
\label{tab:real_results}
\vspace{-1.em}
\end{wraptable}
We further evaluate \ours on real robot platforms to test whether human-derived interaction priors can be deployed without real-world robot data in the target task contexts (\textbf{H3}) and transferred across substantially different embodiments (\textbf{H4}). Additional snapshots and videos are provided in the appendix and supplementary material.

\textbf{\ours solves real-world tasks without real-world robot data.}
On the Franka tabletop setup, we evaluate \ours on \textsc{PourInBowl}, \textsc{HangMugTree}, and \textsc{SortUtensils} (Tab.~\ref{tab:real_results}). These tasks require precise object interactions, grasp-conditioned execution, and robustness to variation in object shape and state. Across ten trials per task, \ours executes the learned interaction motions in real scenes without real-world robot demonstrations, supporting \textbf{H3}.

\textbf{\ours transfers across diverse robot embodiments.}
We further deploy \ours on the RBY-1 humanoid mobile robot for \textsc{PourCup} and \textsc{UseRoller}. Compared with the Franka setup, RBY-1 has substantially different kinematics. \ours achieves success rates of $0.6$ and $0.7$ on the two tasks, respectively, showing that the learned interaction priors are not tied to a specific robot morphology. By resolving embodiment-specific constraints at test time, \ours adapts the same object--object interaction knowledge across diverse robot platforms, supporting \textbf{H4}.

\subsection{Ablation Study: Dissecting Key Components}

\begin{figure}[t]
    \centering

    \begin{subfigure}[t]{0.48\linewidth}
        \centering
        \includegraphics[width=\linewidth]{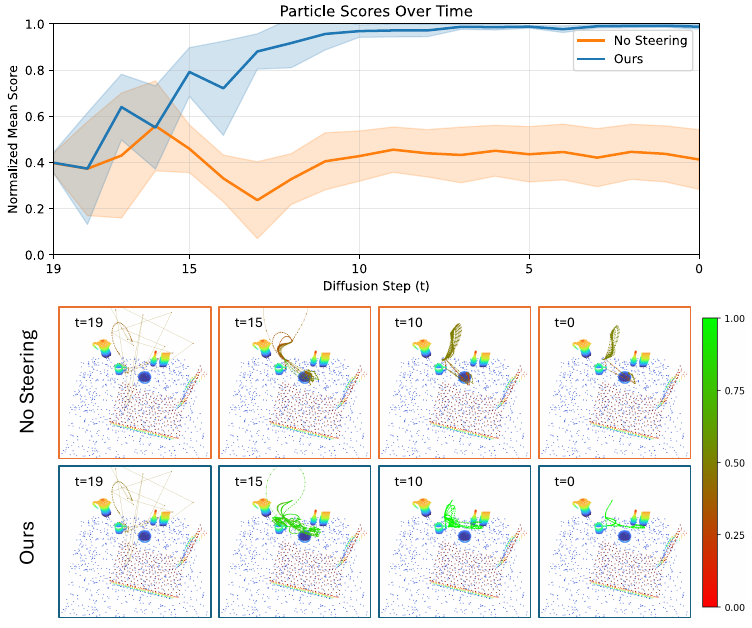}
        \caption{\textbf{Steering visualization.}}
        \label{fig:steer_viz_sub}
    \end{subfigure}
    \hfill
    \begin{subfigure}[t]{0.48\linewidth}
        \centering
        \includegraphics[width=\linewidth]{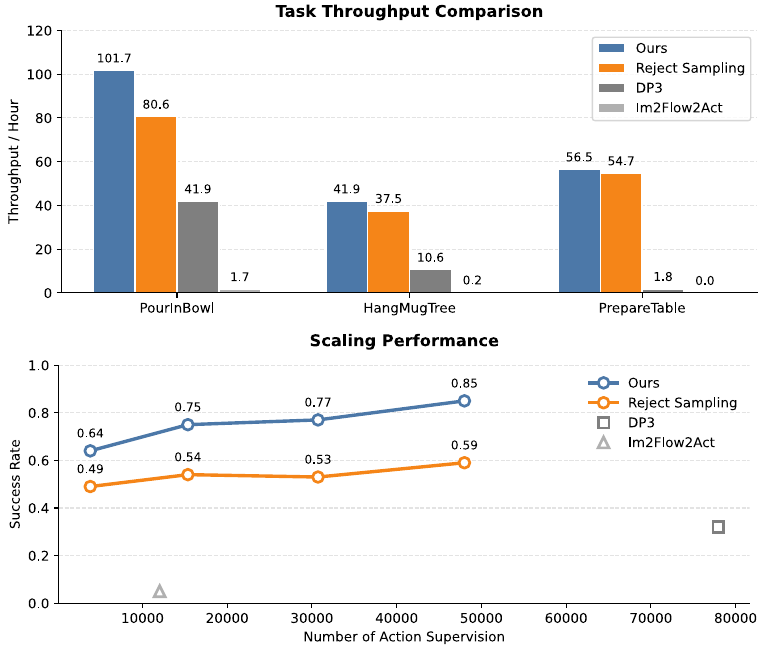}
        \caption{\textbf{Throughput and performance scaling.}}
        \label{fig:throughput_sub}
    \end{subfigure}
    \caption{\textbf{Steering, efficiency, and scaling.}
Left: particle-filtering-based steering guides sampled compositions toward higher-feasibility regions under grasp, kinematic, and scene constraints. Right: \ours improves task-level throughput and achieves better performance with more interaction-motion training data.}
    \label{fig:throughput_steering}
\vspace{-0.75em}
\end{figure}

\textbf{Steering improves generation quality and efficiency. (H5)}
As shown in Fig.~\ref{fig:steer_viz_sub}, steering reallocates particles from low-feasibility regions toward samples that better satisfy grasp, kinematic, collision, and scene constraints during denoising. Qualitative snapshots show assemblies becoming increasingly feasible, with high-quality samples in green and low-quality samples in red. At the task level, \ours achieves the highest throughput across all simulation domains, measured as successful trials within a one-hour budget including inference, planning, and execution (Fig.~\ref{fig:throughput_sub}). This shows that steering improves both motion quality and sampling efficiency by reducing computation spent on infeasible candidates.

\textbf{Performance improves with interaction-motion data scale. (H6)}
We study data scaling on \textsc{PourInBowl} by varying the quantity and diversity of object-centric interaction trajectories (Fig.~\ref{fig:throughput_sub}). Performance improves with richer motion data, suggesting that \ours can effectively leverage diverse interaction supervision. Since these trajectories share the same form as motion extracted from human videos, this provides indirect evidence for scalability with larger human-video sources.

\section{Limitations}

\ours focuses on prehensile manipulation and assumes a rigid attachment between the end-effector and tool object, allowing robot control targets to be recovered from a fixed grasp pose and object--object interaction motion. Relaxing this assumption with learned dynamics models could enable broader non-prehensile skills such as in-hand dexterous manipulation. \ours assumes that the task skeleton is provided; integrating it with task planning or language-guided decomposition would yield a more complete task-and-motion planning system. The current implementation lacks online failure detection and replanning, which would improve robustness under perception noise, object slippage, and unexpected scene changes, especially in long-horizon execution.

\section{Conclusion}
We presented \ours, a framework for transferring interaction-centric manipulation knowledge from human videos to robots without real-world robot demonstrations. By representing manipulation as object-centric motion, \ours separates embodiment-agnostic interaction learning from robot-specific execution through in-context feasibility reasoning and motion planning. Experiments in simulation and on real Franka and RBY-1 platforms show that \ours can solve fine-grained and long-horizon tasks while generalizing across object shapes, scene layouts, and robot embodiments.
%



\bibliographystyle{plainnat}
\bibliography{references}

\clearpage
\newpage

\appendix

\clearpage
\section*{Supplementary Material}
\addcontentsline{toc}{section}{Supplementary Material}

The supplementary material has the following contents:
\begin{itemize}
    \setlength\itemsep{0.15em}
    \setlength\parskip{0pt}
    \item \textbf{Hardware Setup} (Sec.~\ref{sec:app:hardware_setup}): Describes camera, perception, planning, and control details for the Franka and RBY-1 real-world platforms;
    \item \textbf{Tasks and Baselines} (Sec.~\ref{sec:app:tasks_baselines}): Details simulation and real-world task semantics, reset ranges, evaluation protocols, and baseline settings;
    \item \textbf{Human Data Processing Pipeline} (Sec.~\ref{sec:app:human_data}): Describes trajectory fitting, filtering, and qualitative visualizations of RGB-D observations, 3D tracks, and recovered object-centric trajectories;
    \item \textbf{Simulation Data Generation} (Sec.~\ref{sec:app:sim_data}): Describes how robot-specific agent--object grasp data are generated in simulation for Franka and RBY-1;
    \item \textbf{Constraint-Aware Steering Algorithm} (Sec.~\ref{sec:app:steering_alg}): Provides pseudocode for particle-filtering-based steering during constraint-aware composition;
    \item \textbf{Model and Training Details} (Sec.~\ref{sec:app:training_details}): Summarizes model architectures, diffusion sampling settings, and training hyperparameters;
    \item \textbf{Additional Results} (Sec.~\ref{sec:app:additional_results}): Shows qualitative real-world and simulation rollouts across tasks, layouts, and robot embodiments.
\end{itemize}

Additional videos and visualizations are available on our project website: \url{https://human2any.github.io/}

\clearpage
\section{Hardware Setup}
\label{sec:app:hardware_setup}

\begin{figure}[!htbp]
  \centering
  \includegraphics[width=\linewidth]{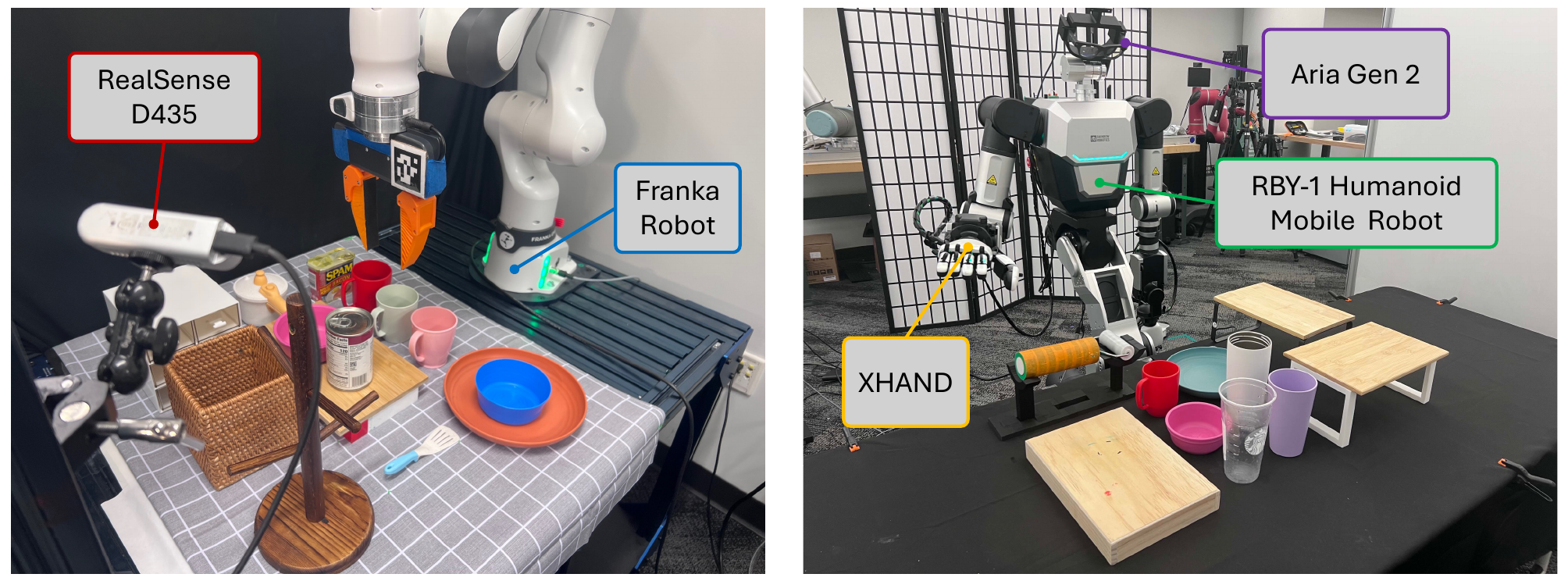}
  \caption{\textbf{Hardware setup.}
  We evaluate on a Franka tabletop setup with an Intel RealSense D435 and an RBY-1 humanoid mobile robot with ROBOTERA XHAND and Meta Aria Gen~2 glasses.}
  \label{fig:hardware}
\end{figure}

Fig.~\ref{fig:hardware} shows the real-world hardware setups used in our experiments. We obtain scene point clouds from calibrated cameras. For the Franka setup, we use an Intel RealSense D435; for the RBY-1 setup, we use Meta Aria Gen~2 glasses and estimate depth maps with FoundationStereo~\cite{wen2025stereo}. We segment objects with SAM~2~\cite{ravi2024sam} and identify task-relevant objects by extracting CLIP~\cite{radford2021learning} features for each mask region. Once a valid robot trajectory is planned, we execute it with platform-specific controllers: OSC~\cite{khatib2003unified} at $20\,\mathrm{Hz}$ for Franka end-effector targets, and joint-position control for the RBY-1 torso, arms, and dexterous hands, with PD control for base tracking.

\clearpage
\section{Tasks and Baselines}
\label{sec:app:tasks_baselines}

\subsection{Task Descriptions}
\label{sec:app:task_descriptions}

We evaluate \ours on simulation and real-world manipulation tasks that require interaction-centric motion generation, constraint-aware composition, and generalization across object shapes, poses, layouts, and robot embodiments. Each task is specified by a high-level task skeleton, while the low-level object interaction motions and robot execution trajectories are generated by \ours.

\paragraph{PourInBowl.}
The robot picks up a container and pours its contents into a target bowl. This task requires generating a precise object--object pouring motion while satisfying grasp, reachability, and collision constraints. We use this task to evaluate fine-grained interaction motion and adaptation to different container shapes, bowl poses, and surrounding layouts.

\paragraph{HangMugTree.}
The robot grasps a mug and hangs it on a mug tree. Unlike tasks with a fixed target pose, the target branch is identified at test time based on robot and scene feasibility, allowing the system to avoid collisions and choose an executable hanging motion. This task evaluates precision manipulation, grasp-conditioned alignment, and constraint-aware target reasoning.

\paragraph{PrepareTable.}
The robot performs a multi-stage table-setting task, such as placing a bowl on a plate and then placing a cookie box on the bowl. This task requires composing multiple interaction skills over a longer horizon. We use this task to evaluate long-horizon skill chaining and compositional generalization across different object arrangements and scene layouts.

\paragraph{SortUtensils.}
In the real Franka setup, the robot places a bowl on a plate and then places a utensil on the bowl, requiring multi-stage interaction composition. This task evaluates long-horizon skill chaining and generalization across different object arrangements and scene layouts.

\paragraph{PourCup.}
On the RBY-1 humanoid mobile robot, the robot performs a cup-pouring task in a larger workspace. This task evaluates whether the object-centric pouring representation can be grounded on a different robot embodiment and workspace.

\paragraph{UseRoller.}
The RBY-1 robot uses a roller-like tool to interact with a target surface or object. This task evaluates tool-use behavior with a dexterous hand and mobile humanoid embodiment, requiring the learned interaction prior to be grounded through embodiment-specific grasping and motion constraints.

\paragraph{Task variants and generalization.}

\begin{figure}[H]
  \centering
  \includegraphics[width=0.8\linewidth]{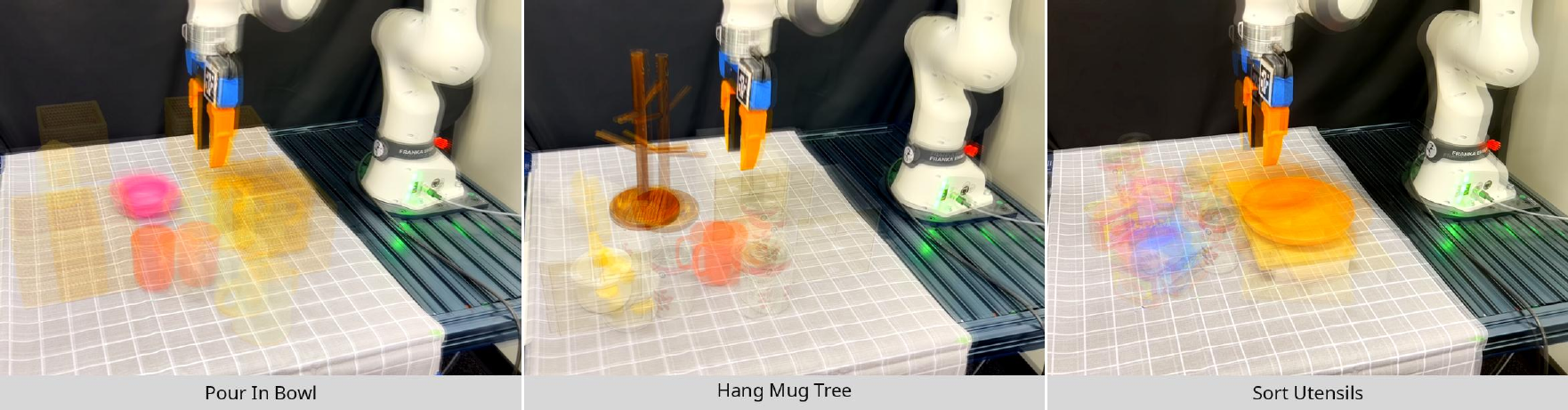}
  \caption{\textbf{Real-world reset settings.}
  Object pose randomization ranges used for Franka real-world evaluation.}
  \label{fig:real_reset}
\end{figure}

\begin{figure}[H]
  \centering
  \includegraphics[width=0.7\linewidth]{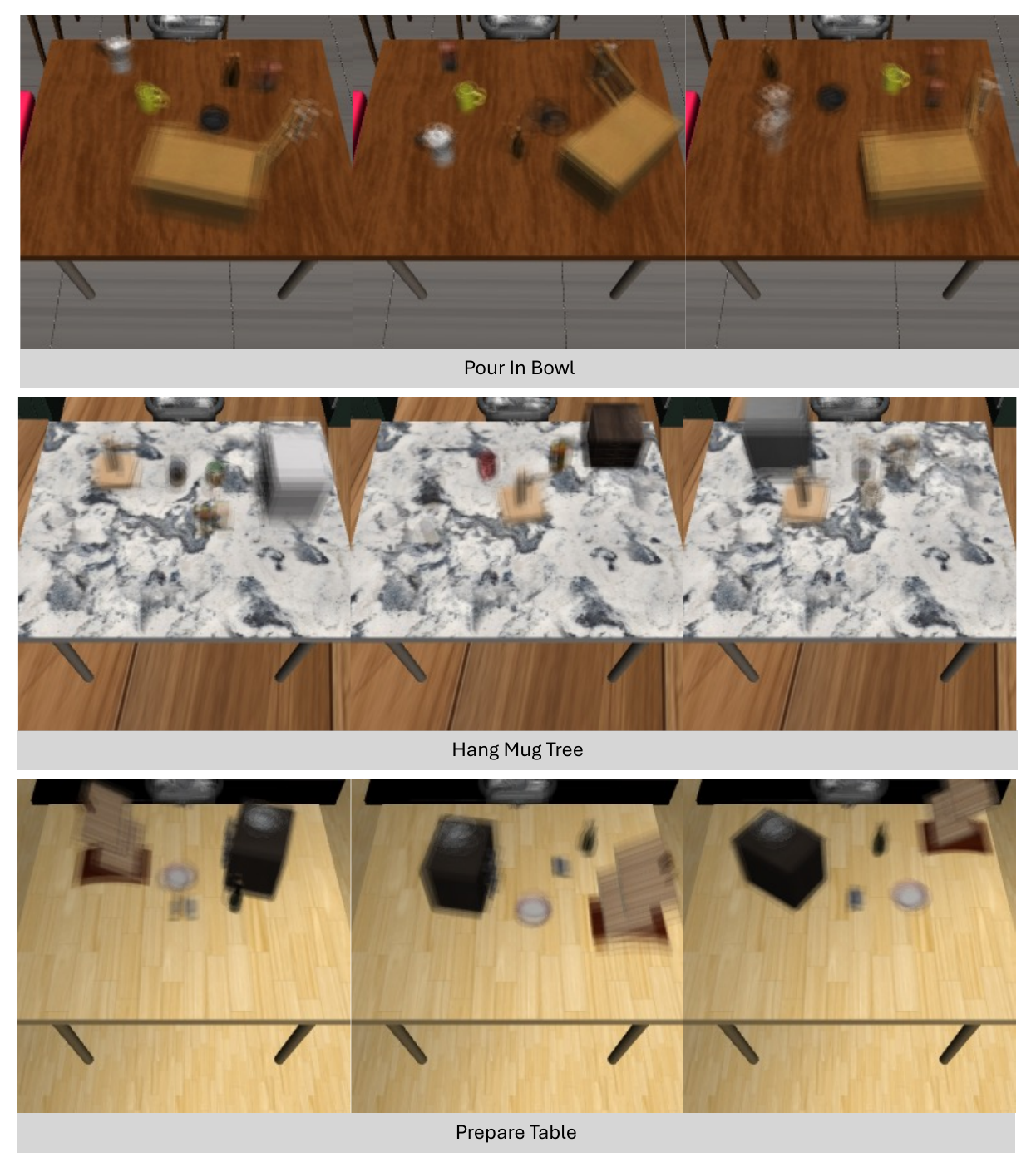}
  \caption{\textbf{Simulation reset settings.}
  Global layout variants and local object pose randomization ranges used in simulation.}
  \label{fig:sim_reset}
\end{figure}

The simulation domains test both local robustness and broader contextual generalization. Each domain contains three variants with distinct global layouts, object arrangements, and interaction contexts; within each variant, object poses are randomized by approximately $0.1$ m in translation and $30^\circ$ in rotation. For each domain, we train on two variants and hold out the third for OOD evaluation, running $50$ trials per variant. In real-world experiments, we vary object shapes and poses across trials and evaluate on both Franka and RBY-1, running $10$ trials per task. The simulation reset ranges and real-world Franka reset settings are illustrated in Fig.~\ref{fig:sim_reset} and Fig.~\ref{fig:real_reset}, respectively.

\subsection{Baseline Settings}

\paragraph{Training data.} We use strong supervised variants of the baselines whenever possible. In simulation, DP3 and Im2Flow2Act are trained with $100$ robot demonstrations per task from MimicLab~\cite{saxena2025mimiclabs}. Although the original Im2Flow2Act trains its flow-conditioned policy from predefined random exploration data, designing such exploration rules is non-trivial for our long-horizon, contact-rich tasks. We therefore provide robot demonstrations for both its flow generation module and flow-conditioned policy, giving the baseline at least as strong supervision as task-agnostic exploration data.

For real-world tasks, DP3 and Im2Flow2Act are trained with $50$ real-robot teleoperation demonstrations per task. Since we do not assume a digital twin, we retrain the Im2Flow2Act flow-conditioned policy on real-world teleoperation data. Its flow generation module is trained with the same $50$ teleoperation demonstrations plus $50$ additional human demonstration videos. For Im2Flow2Act, we downsample the generated flow by $n=6$ in simulation to fit GPU memory; we also tested accumulated delta end-effector actions, but this did not improve performance.

\paragraph{Baseline performance analysis.} The low performance of the baselines highlights the importance of constraint-aware composition. DP3 and Im2Flow2Act receive strong supervision, but do not explicitly reason over how predicted motions satisfy the current grasp, kinematic, collision, and scene constraints when composed into long-horizon executions. For Im2Flow2Act, even when the predicted flow represents plausible object motion, translating it into robot actions remains constraint-dependent: not every flow can be realized under the robot's current grasp, workspace, and collision constraints. Thus, locally plausible predictions can still become infeasible under diverse layouts and object poses. In contrast, \ours composes interaction and grasp motions under in-context feasibility constraints, enabling executable rather than merely plausible motion combinations.

\clearpage
\section{Human Data Processing Pipeline}
\label{sec:app:human_data}

\subsection{Trajectory Fitting and Filtering}
\label{sec:app:traj_fitting}

Given RGB-D human demonstration videos, we first segment the tool and target objects and track 3D points on the tool object across time~\cite{xiao2025spatialtrackerv2}. The tracked points are transformed into a common world frame (robot base frame) using calibrated camera poses. We then recover the object-centric motion by fitting a rigid transform from the first-frame tool geometry to each subsequent frame. To reduce the effect of noisy or drifting tracks, we use RANSAC~\cite{fischler1981random} to select inlier correspondences before applying Kabsch alignment~\cite{lawrence2019purely}. This produces a raw relative trajectory $\tau_{\mathrm{raw}}^{\mathrm{rel}}$.

We further refine the trajectory with temporal smoothing. Specifically, we optimize the per-frame rotations and translations to fit the tracked 3D points while penalizing abrupt changes between consecutive frames. This produces a smoothed relative trajectory $\tau_{\mathrm{smooth}}^{\mathrm{rel}}$, which is used as the object-centric interaction motion for training the object--object interaction sampler. Alg.~\ref{alg:traj_extraction} summarizes the full fitting and filtering procedure, and Alg.~\ref{alg:ransac_inliers} details the RANSAC-based inlier selection.

\begin{algorithm}[H]
\small
\caption{Object-Centric Trajectory Extraction from Human Video Tracks}
\label{alg:traj_extraction}
\begin{algorithmic}[1]
\Require Tracked 3D points in camera frame $\mathbf{P}^{\mathrm{cam}}\in\mathbb{R}^{H\times M\times 3}$, camera-to-world transform $\mathbf{T}^{\mathrm{world}}_{\mathrm{cam}}$
\Ensure Raw and smoothed relative object trajectories $\tau_{\mathrm{raw}}^{\mathrm{rel}}, \tau_{\mathrm{smooth}}^{\mathrm{rel}}$

\State Transform tracked points to the world frame:
$\mathbf{P}\leftarrow \mathbf{T}^{\mathrm{world}}_{\mathrm{cam}}\mathbf{P}^{\mathrm{cam}}$
\State Use first-frame points $\mathbf{P}_0=\{\mathbf{p}_{0,m}\}_{m=1}^{M}$ as the canonical object geometry
\State Initialize empty raw trajectory $\tau_{\mathrm{raw}}^{\mathrm{rel}}$

\For{$t=0,\ldots,H-1$}
    \State $\mathcal{I}_t \leftarrow \Call{RansacInliers}{\mathbf{P}_0,\mathbf{P}_t}$
    \State Estimate rigid transform $(\mathbf{R}_t,\mathbf{t}_t)$ from $\{(\mathbf{p}_{0,m},\mathbf{p}_{t,m})\}_{m\in\mathcal{I}_t}$ using Kabsch alignment~\cite{lawrence2019purely}
    \State Append
    $\mathbf{T}^{\mathrm{raw}}_t=
    \begin{bmatrix}
    \mathbf{R}_t & \mathbf{t}_t\\
    \mathbf{0}^{\top} & 1
    \end{bmatrix}$
    to $\tau_{\mathrm{raw}}^{\mathrm{rel}}$
\EndFor

\State Refine $\tau_{\mathrm{raw}}^{\mathrm{rel}}$ by optimizing rotations $\{\mathbf{R}_t\}_{t=0}^{H-1}$ and translations $\{\mathbf{t}_t\}_{t=0}^{H-1}$:
\[
\min_{\{\mathbf{R}_t,\mathbf{t}_t\}_{t=0}^{H-1}}
\sum_{t=0}^{H-1}\sum_{m=1}^{M}
\left\|
\mathbf{R}_t\mathbf{p}_{0,m}+\mathbf{t}_t-\mathbf{p}_{t,m}
\right\|_2^2
+
\lambda_R\sum_{t=1}^{H-1} d_R(\mathbf{R}_t,\mathbf{R}_{t-1})^2
+
\lambda_T\sum_{t=1}^{H-1}
\|\mathbf{t}_t-\mathbf{t}_{t-1}\|_2^2 .
\]
\Statex \hspace{\algorithmicindent}Here, $d_R(\cdot,\cdot)$ is the rotation distance, and $\lambda_R,\lambda_T$ control temporal smoothness.
\State Convert the optimized poses into $\tau_{\mathrm{smooth}}^{\mathrm{rel}}=\{\mathbf{T}^{\mathrm{smooth}}_t\}_{t=0}^{H-1}$
\State \Return $\tau_{\mathrm{raw}}^{\mathrm{rel}}$, $\tau_{\mathrm{smooth}}^{\mathrm{rel}}$
\end{algorithmic}
\end{algorithm}

\begin{algorithm}[H]
\small
\caption{RANSAC Inlier Selection for Rigid Alignment}
\label{alg:ransac_inliers}
\begin{algorithmic}[1]
\Require Corresponding point sets $\mathbf{P}_0=\{\mathbf{p}_{0,m}\}_{m=1}^{M}$ and $\mathbf{P}_t=\{\mathbf{p}_{t,m}\}_{m=1}^{M}$, batch size $B$, number of iterations $N_{\mathrm{iter}}$, inlier threshold $\epsilon$
\Ensure Inlier set $\mathcal{I}^{\star}$

\State Initialize best inlier set $\mathcal{I}^{\star}\leftarrow \emptyset$
\For{$j=1,\ldots,N_{\mathrm{iter}}$}
    \State Randomly sample an index subset $\mathcal{B}_j\subseteq\{1,\ldots,M\}$ with $|\mathcal{B}_j|=B$
    \State Estimate $(\mathbf{R}_j,\mathbf{t}_j)$ from $\{(\mathbf{p}_{0,m},\mathbf{p}_{t,m})\}_{m\in\mathcal{B}_j}$ using Kabsch alignment~\cite{lawrence2019purely}
    \State Compute residuals for all correspondences:
    $e_m=\|\mathbf{R}_j\mathbf{p}_{0,m}+\mathbf{t}_j-\mathbf{p}_{t,m}\|_2$
    \State Set $\mathcal{I}_j=\{m\mid e_m<\epsilon\}$
    \If{$|\mathcal{I}_j|>|\mathcal{I}^{\star}|$}
        \State $\mathcal{I}^{\star}\leftarrow \mathcal{I}_j$
    \EndIf
\EndFor
\State \Return $\mathcal{I}^{\star}$
\end{algorithmic}
\end{algorithm}

\subsection{Data Visualization and Analysis}
\label{sec:app:data_viz}

\begin{figure}[!htbp]
  \centering
  \includegraphics[width=\linewidth]{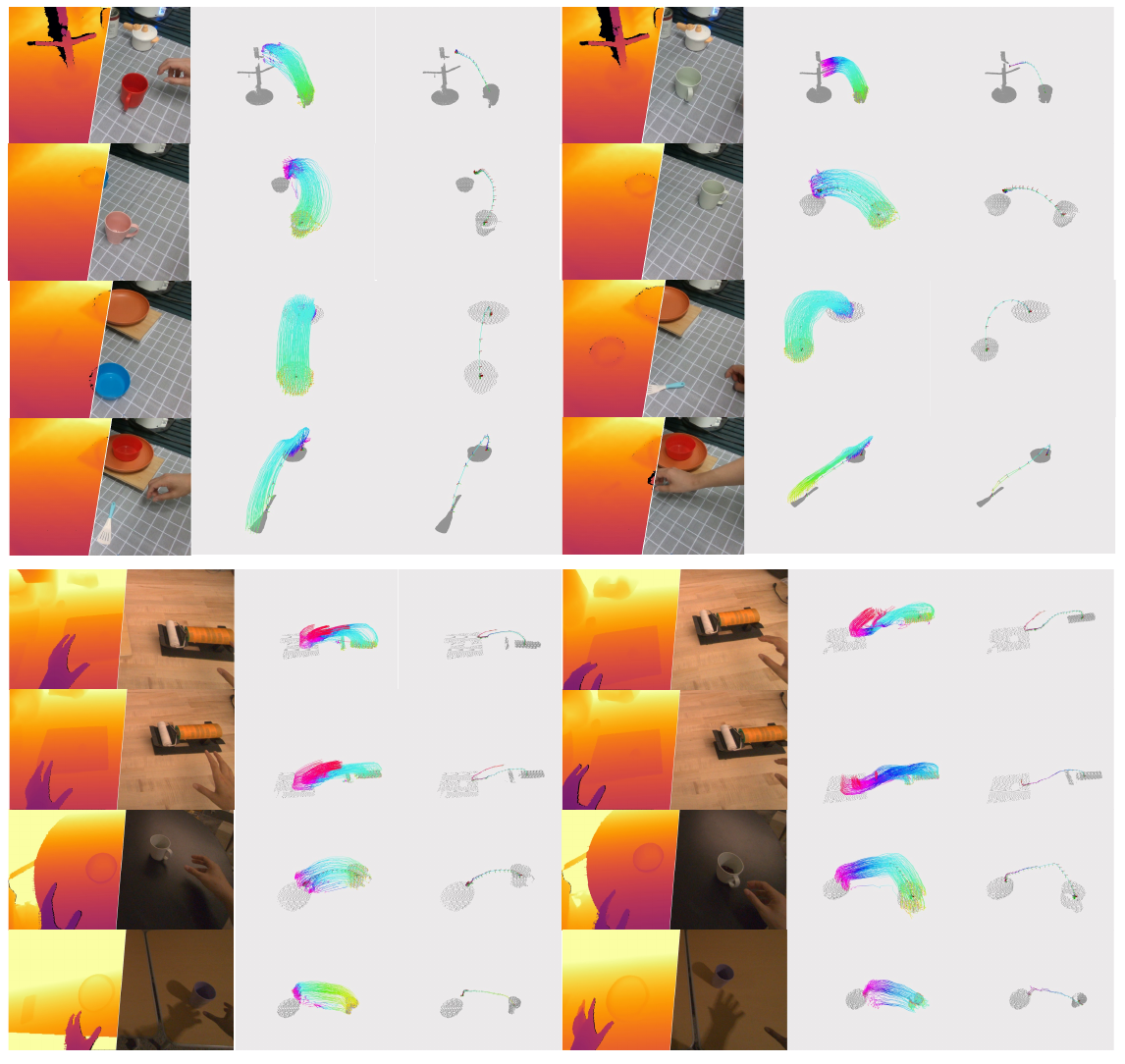}
  \caption{\textbf{Human data processing.}
  We visualize RGB observations with depth overlays, 3D point tracks, and recovered object-centric trajectories from human demonstrations.}
  \label{fig:human_data}
\end{figure}

Fig.~\ref{fig:human_data} visualizes intermediate outputs from the human data processing pipeline. We show RGB observations with depth overlays, tracked 3D points, and the recovered object-centric trajectories. The visualizations illustrate that the pipeline extracts consistent object motion across diverse demonstrations and filters noisy tracks into smooth relative trajectories suitable for training object--object interaction samplers.

\clearpage
\section{Simulation Data Generation}
\label{sec:app:sim_data}

We generate robot-specific agent--object training data in simulation. For the Franka setup, we sample grasp candidates in the object local frame using geometric heuristics, execute the corresponding grasp motions in simulation, and retain successful executions as positive samples, following prior work~\cite{acronym2020,murali2025graspgen}. We further augment the dataset by applying random rigid transformations to both the object and associated grasp trajectories, improving robustness to object pose variation.

For the RBY-1 dexterous hand, we use DRO~\cite{wei2024dro} to generate candidate dexterous grasps. We curate the dataset by filtering out candidates that can stably hold the object in the air but still contain undesirable collisions or penetrations between the robot hand and the object. The resulting simulation datasets provide embodiment-specific agent--object priors for both the Franka gripper and the RBY-1 dexterous hand, without requiring real-world robot demonstrations.

\clearpage
\section{Constraint-Aware Steering Algorithm}
\label{sec:app:steering_alg}

Alg.~\ref{alg:pf_ddpm} summarizes the particle-filtering-based steering procedure used for constraint-aware composition. During joint denoising, candidate interaction and grasp motions are assembled into robot trajectories, scored by in-context feasibility checks, and resampled toward executable compositions.

\begin{algorithm}[H]
\footnotesize
\caption{Particle-Filter-Based Guided Joint Diffusion Sampling}
\label{alg:pf_ddpm}
\begin{algorithmic}[1]
\Require test-time context $c$; segment-level conditioning $\tilde{c}^{1:K}$; pretrained diffusion models $\{\epsilon_{\theta_k}\}_{k=1}^{K}$; assembly operator $\Gamma$; constraint score $\mathcal{S}$; temperature $\beta$; number of particles $M$; diffusion steps $T$
\Ensure feasible compositional motion $\tau^\star$ 

\State Sample initial particles $\mathbf{x}_{T,1:M}^{k} \sim \mathcal{N}(\mathbf{0}, \mathbf{I})$, $\forall k \in \{1,\dots,K\}$
\For{$t=T,\dots,1$}
        \For{$k=1,\dots,K$}
            \State $\hat{\mathbf{x}}_{0,1:M}^{\,k}
            \gets
            \frac{1}{\sqrt{\bar{\alpha}_t}}
            \left(
            \mathbf{x}_{t,1:M}^{k}
            -
            \sqrt{1-\bar{\alpha}_t}\,
            \epsilon_{\theta_k}(\mathbf{x}_{t,1:M}^{k}, t, \tilde{c}^{k})
            \right)$ \Comment{Compute denoised estimate}
        \EndFor
        \State $\tau_{i} \gets \Gamma(\hat{\mathbf{x}}_{0,i}^{\,1:K}, c), \forall i \in \{1,\dots,M\}$ \Comment{Assemble trajectory}
        \State ${w}_{t,i} \gets \exp\!\big(\beta\,\mathcal{S}(\tau_i, c)\big), \forall i \in \{1,\dots,M\}$ \Comment{Evaluate constraints to compute weights}
    \State Normalize ${w}_{t,1:M}$ to obtain $\tilde{w}_{t,1:M}$
    \State Resample $\{\mathbf{\tilde{x}}_{t,i}^{1:K}\}_{i=1}^{M}$ according to $\tilde{w}_{t,1:M}$
            \For{$k=1,\dots,K$}
                \State $\mathbf{{x}}_{t-1,1:M}^{k} \sim p_{\theta_k}(\mathbf{x}_{t-1, 1:M}^{k}\mid \mathbf{\tilde{x}}_{t,1:M}^{k}, \tilde{c}^{k})$ \Comment{Sample reverse step}
            \EndFor
\EndFor
\State $i^\star \gets \arg\max_i w_{1,i}$
\State $\tau^\star = \Gamma(\mathbf{x}_{0,i^\star}^{1:K}, c)$
\State \Return $\tau^\star$
\end{algorithmic}
\end{algorithm}

\clearpage
\section{Model and Training Details}
\label{sec:app:training_details}

We implement all trajectory samplers using the CleanDiffuser codebase~\cite{cleandiffuser}. Each sampler is trained as a conditional DDPM over pose trajectories. The model predicts $6$-DoF pose actions with action dimension $6$ and horizon $H=16$, conditioned on point-cloud observations. We use a point-transformer condition encoder with feature dimension $256$ and a U-Net structure~\cite{Ze2024DP3} for denoising. During sampling, we use $20$ reverse diffusion steps. All models are trained with batch size $512$ for $60{,}000$ gradient steps using Adam with learning rate $10^{-4}$.

\clearpage
\section{Additional Results}
\label{sec:app:additional_results}

\subsection{Real-world Rollouts}
\label{sec:app:real_rollouts}

\begin{figure}[!htbp]
  \centering
  \includegraphics[width=\linewidth]{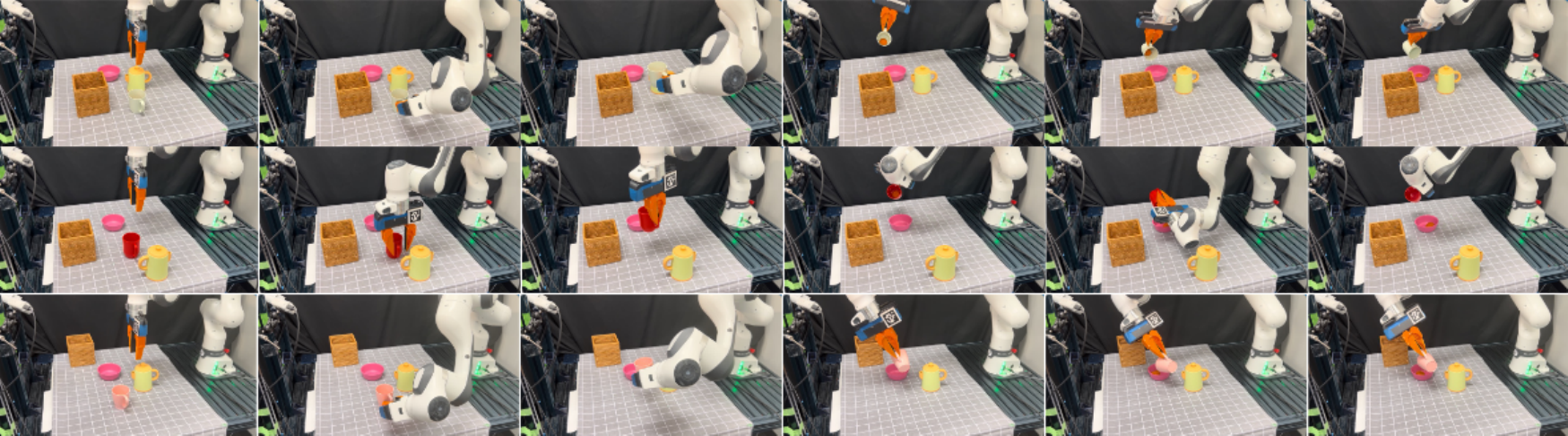}
  \caption{\textbf{Real-world rollouts on PourInBowl tasks.}
  The robot avoids the kettle, grasps the cup by its handle, and rotates it during the transfer motion to align the cup for pouring into the bowl.}
  \label{fig:real_franka_pou}
\end{figure}

\begin{figure}[!htbp]
  \centering
  \includegraphics[width=\linewidth]{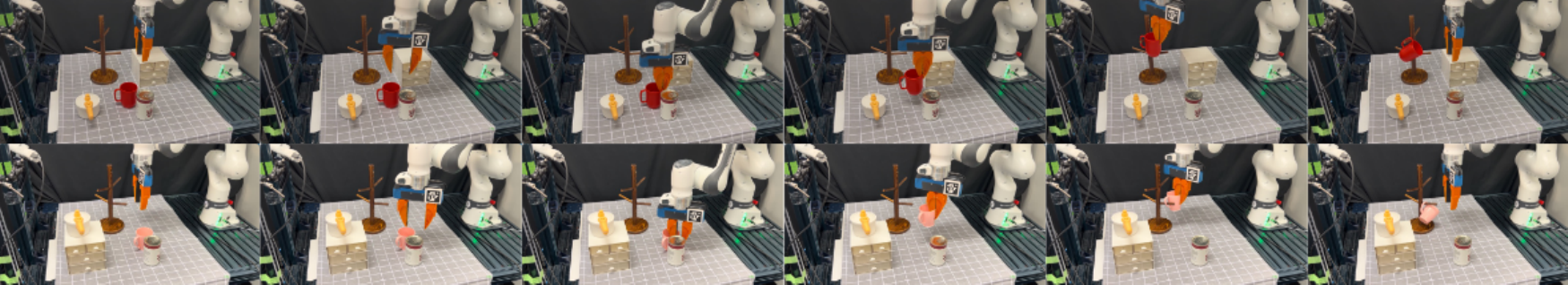}
  \caption{\textbf{Real-world rollouts on HangMugTree tasks.}
  The robot aligns its grasp to avoid non-target objects and reorients the mug on the fly to accurately place the handle onto the mug tree.}
  \label{fig:real_franka_hangmugtree}
\end{figure}

\begin{figure}[!htbp]
  \centering
  \includegraphics[width=\linewidth]{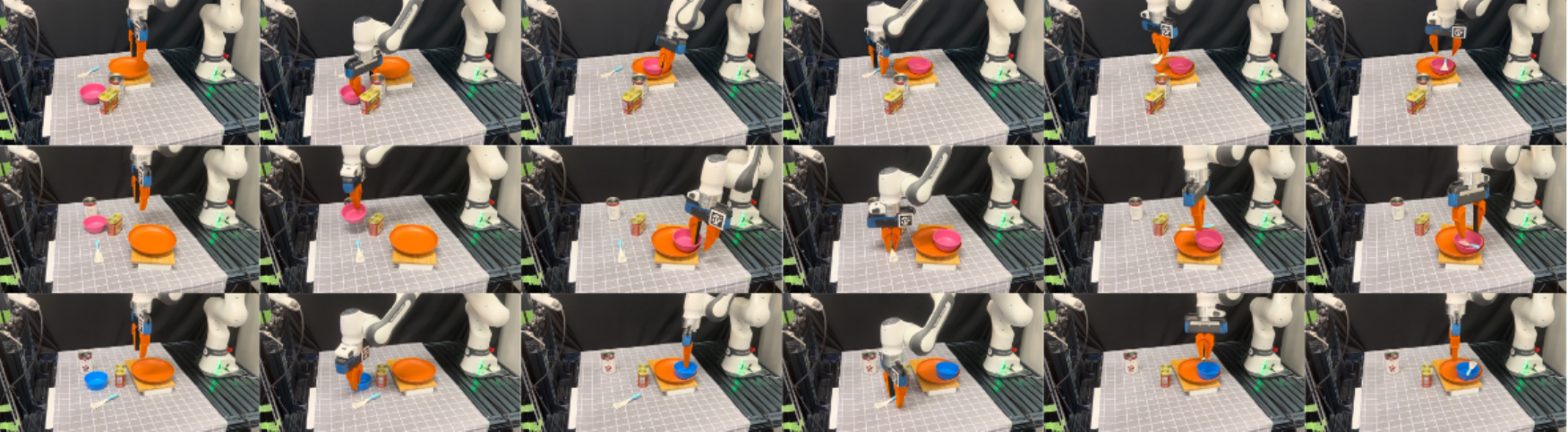}
  \caption{\textbf{Real-world rollouts on {SortUtensils} tasks.} 
  The robot sequentially places a bowl onto a plate and a utensil onto the bowl, demonstrating long-horizon skill composition across varying object arrangements and scene layouts.
   }
  \label{fig:real_franka_serve}
\end{figure}

\begin{figure}[!htbp]
  \centering
  \includegraphics[width=\linewidth]{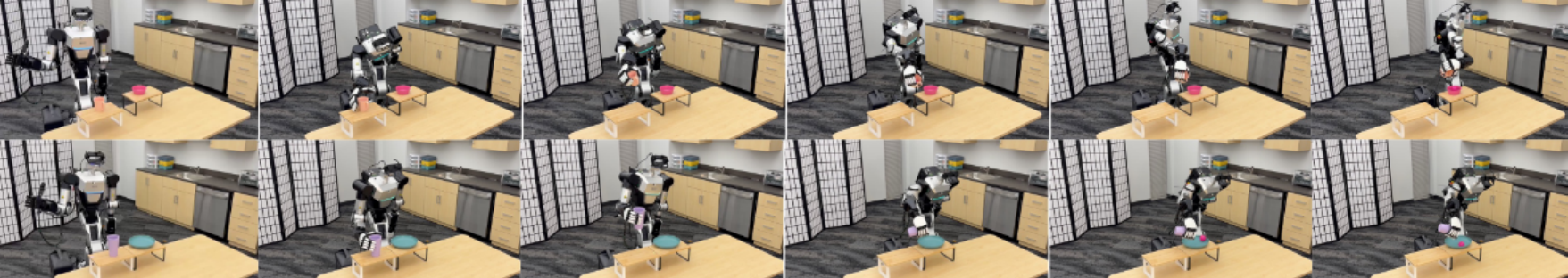}
  \caption{\textbf{Real-world rollouts on PourCup tasks.}
The robot firmly grasps the cup and leverages whole-body motion to pour the pink ball onto the target container.}
  \label{fig:real_rby_pour}
\end{figure}

\begin{figure}[!htbp]
  \centering
  \includegraphics[width=\linewidth]{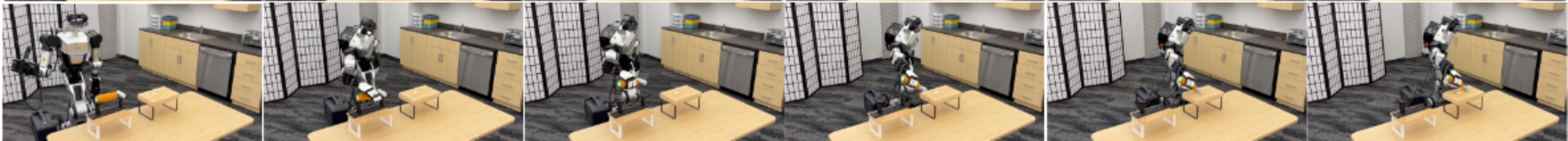}
  \caption{\textbf{Real-world rollouts on UseRoller tasks.} The robot grasps the handle of the roller precisely and move to press the target dough.}
  \label{fig:real_rby_roller}
\end{figure}

Fig.~\ref{fig:real_franka_pou} - Fig.~\ref{fig:real_rby_roller} show representative real-world executions of \ours on both the Franka tabletop setup and the RBY-1 humanoid mobile robot. These rollouts cover interaction-centric skills such as pouring, hanging, table setting, and tool use, demonstrating that human-derived object--object interaction priors can be composed with robot-specific constraints to produce executable motions across different embodiments and scene contexts.


\subsection{Simulation Rollouts}
\label{sec:app:sim_rollouts}

\begin{figure}[!htbp]
  \centering
  \includegraphics[width=\linewidth]{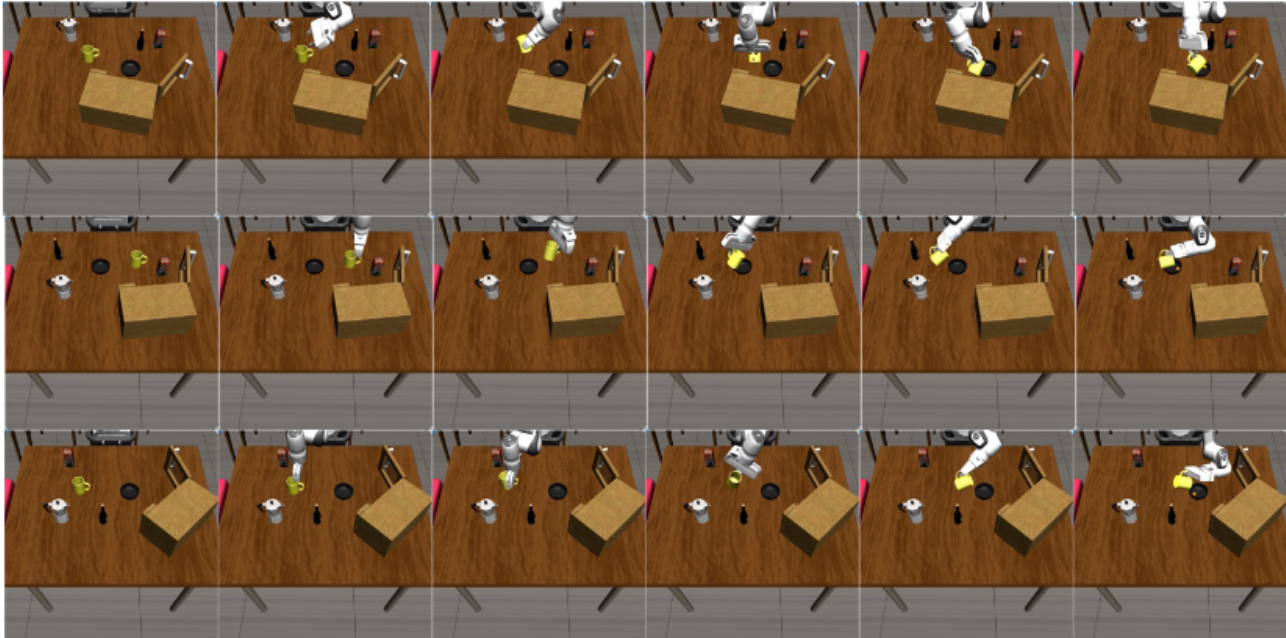}
  \caption{\textbf{Simulation rollouts.}
Qualitative \textsc{PourInBowl} executions showing precise pouring interactions under varied object poses and scene layouts.}
  \label{fig:sim_task3}
  \vspace{-10pt}
\end{figure}

\begin{figure}[!htbp]
  \centering
  \includegraphics[width=\linewidth]{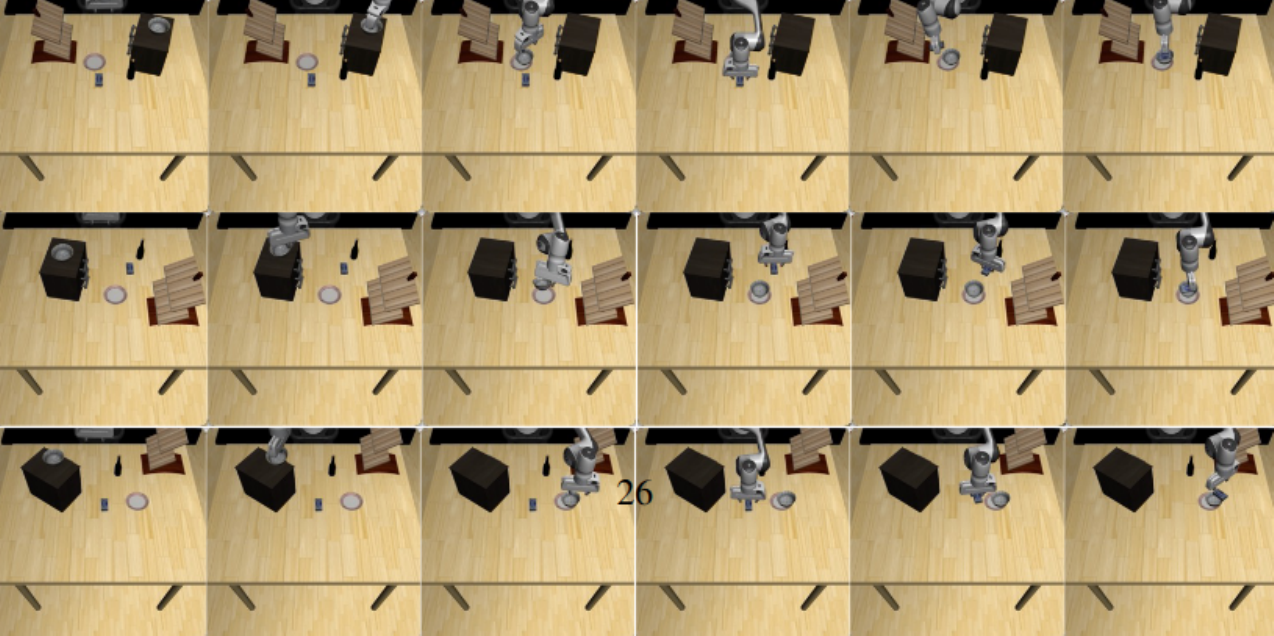}
 \caption{\textbf{Simulation rollouts.}
Qualitative executions of \ours on \textsc{PrepareTable}, demonstrating long-horizon skill chaining through multi-stage interaction composition.}
  \label{fig:sim_task1}
  \vspace{-10pt}
\end{figure}

Fig.~\ref{fig:sim_task1} - Fig.~\ref{fig:sim_task3} show representative simulation executions across our task domains. These rollouts illustrate how \ours composes learned interaction priors with robot- and scene-specific constraints to generate executable trajectories for fine-grained manipulation, long-horizon skill chaining, and diverse object arrangements.

\begin{figure}[!htbp]
  \centering
  \includegraphics[width=\linewidth]{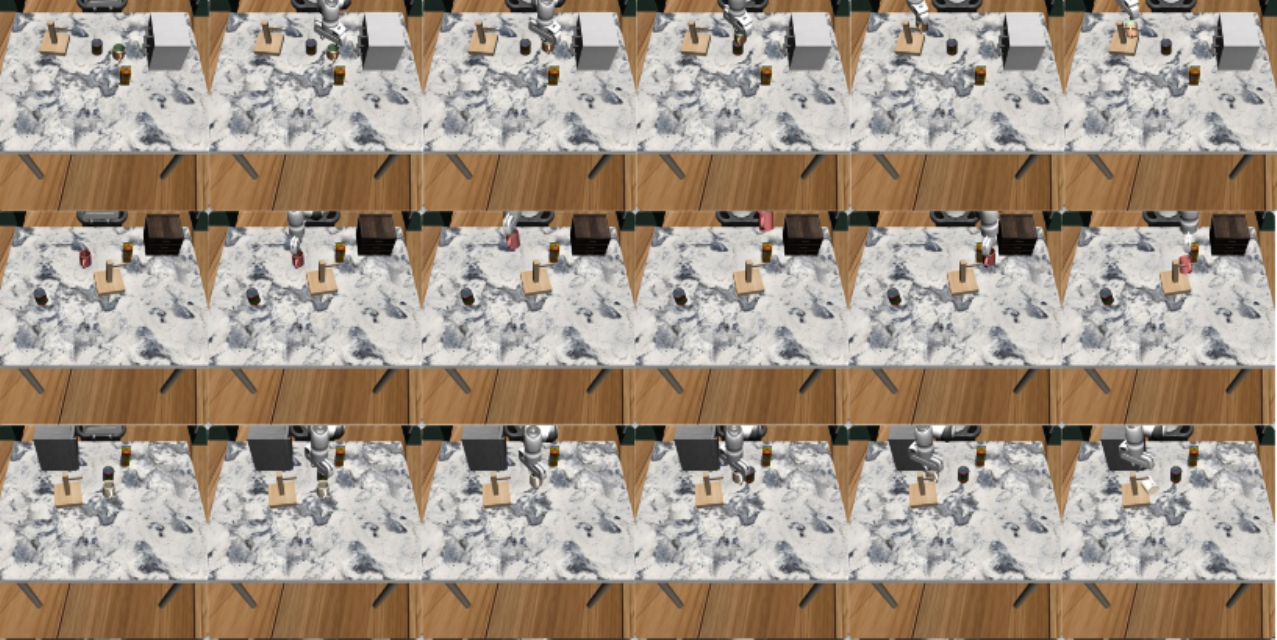}
  \caption{\textbf{Simulation rollouts.}
Qualitative \textsc{HangMugTree} executions showing fine-grained interaction motion and precise mug--branch alignment.}
  \label{fig:sim_task2}
  \vspace{-10pt}
\end{figure}

\end{document}